\documentclass{article}
\usepackage{amsmath,graphicx}
\usepackage{color}
\usepackage{booktabs}
\usepackage{amsfonts}
\usepackage{slashbox}
\usepackage{subfig}
\usepackage{hyperref}
\usepackage{float}
\usepackage[preprint]{spconf}

\title{Sequence-based Multi-lingual Low Resource Speech Recognition}
\name{Siddharth Dalmia, Ramon Sanabria, Florian Metze and Alan W. Black}
\address{Language Technologies Institute, Carnegie Mellon University; Pittsburgh, PA; U.S.A.\\
\texttt{\{sdalmia|ramons|fmetze|awb\}@cs.cmu.edu}}
\begin{document}
\ninept
\copyrightnotice{\copyright\ IEEE 2018}
\toappear{To appear in {\it Proc.\ ICASSP 2018, April 15-20, 2018, Calgary, Canada}}

\maketitle
\begin{abstract}
Techniques for multi-lingual and cross-lingual speech recognition can help in low resource scenarios, to bootstrap systems and enable analysis of new languages and domains. End-to-end approaches, in particular sequence-based techniques, 
are attractive because of their simplicity and elegance.
While it is possible to integrate traditional multi-lingual bottleneck feature extractors as front-ends, we show that end-to-end multi-lingual training of sequence models is effective on context independent models trained using Connectionist Temporal Classification (CTC) loss. We show that our model improves performance on Babel languages by over 6\% absolute in terms of word/phoneme error rate when compared to mono-lingual systems built in the same setting for these languages. We also show that the trained model can be adapted cross-lingually to an unseen language using just 25\% of the target data. 
We show that training on multiple languages is important for very low resource cross-lingual target scenarios, but not for multi-lingual testing scenarios. Here, it appears beneficial to include large well prepared datasets.
\end{abstract}
\begin{keywords}
multi-lingual speech recognition, cross-lingual adaptation, connectionist temporal classification, feature representation learning
\end{keywords}

\section{Introduction}
\label{sec:intro}
State-of-the-art speech recognition systems with human-like performance~\cite{saon2017english,xiong2016achieving} are trained on hundreds of hours of well-annotated speech. Since annotation is an expensive and time-consuming task, similar performance is typically unattainable on low resource languages. Multi-lingual or cross-lingual techniques allow transfer of models or features from well-trained scenarios to those where large amounts of training data may not be available, cannot be transcribed, or are otherwise hard to come by~\cite{stolcke2006cross,grezl2016study}.


The standard approach is to train a context dependent Hidden Markov Model based Deep Neural Network acoustic model with a ``bottleneck'' layer using a frame based criterion on a large multi-lingual corpus~\cite{vesely2012language,knill2013investigation,thang:sltu2012}. The network up to the bottleneck layer can be used as a language-independent feature extractor while adapting to a new language. Generating such a model requires the preparation of frame level segmentation in each language, which is usually achieved by training separate mono-lingual systems first. This is a cumbersome multi-step process. Moreover, if the speaking style, acoustic quality, or linguistic properties of the recordings are very different across a set of languages, the segmentations may be inconsistent across languages and thus sub-optimal for generating features in a new language.


\begin{figure}[hb!]
  \centering
  \includegraphics[width=\linewidth]{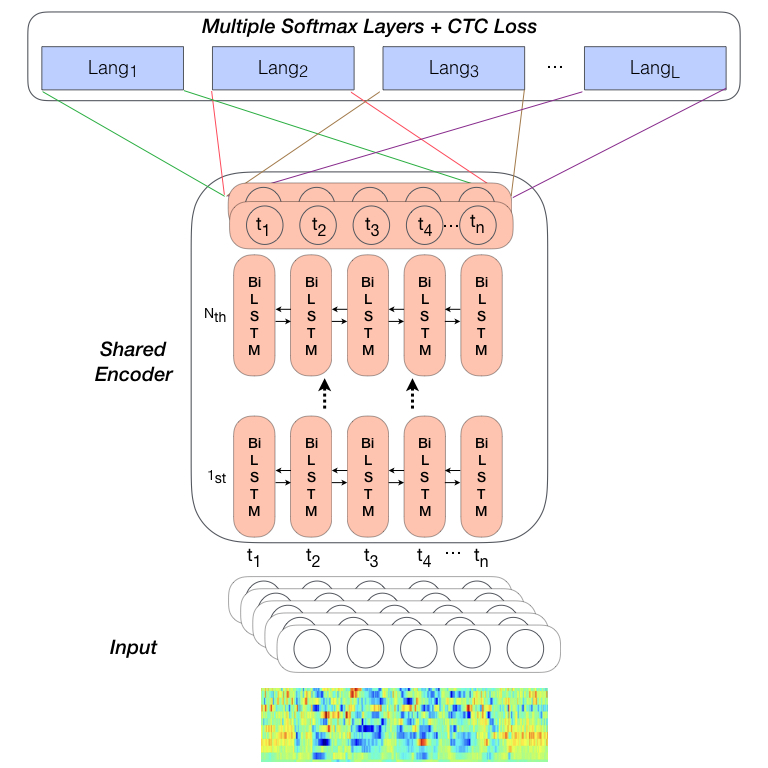}
  \caption{Multi-lingual CTC model following the ``shared hidden layer'' approach for LSTM layers.}
  \label{fig:speech_production}
\end{figure}

On the other hand, end-to-end training approaches which directly model context independent phones are elegant, and greatly facilitate speech recognition training. Most do not require an explicit alignment of transcriptions with the training data, and there are typically fewer hyper-parameters to tune.
We show that sequence training in multi-lingual settings can create feature extractors, which can directly be ported to new languages using a linear transformation (on very limited data), or re-trained on more data, opening a door to end-to-end language universal speech recognition. 

\section{Related Work and Babel Dataset}
\label{sec:related}



Some of the early works in multi-lingual and cross-lingual speech recognition involved the use of language independent features like articulatory features~\cite{stuker2003integrating} to train HMM based systems. Authors in~\cite{burget2010multilingual} used subspace Gaussian mixture model to map phonemes of different languages together. Authors in~\cite{schultz2001language} introduce the use of a shared phone set to build HMM based language independent acoustic models and show the adaptation of pre-existing models towards a new language.\\
With the on-set of deep learning the focus of the models shifted to learning features across languages which can be mapped to the same space~\cite{stolcke2006cross,ghoshal2013multilingual}. Authors in~\cite{swietojanski2012unsupervised} looked at unsupervised pretraining on different languages for a cross lingual recognition. The dominant architecture for multi-lingual or cross-lingual speech recognition has been the so-called ``shared hidden layer'' model, in which data is passed through a series of shared feed-forward layers, before being separated into multiple language-specific softmax layers, which are trained using cross-entropy~\cite{scanzio2008use,vesely2012language,heigold2013multilingual}. This architecture can also be used as a ``bottleneck'' feature extractor, from which ``language independent'' features are extracted, on top of which a target-language acoustic model can be built. Authors in~\cite{grezl2014adaptation} showed that these multi-lingual models can be adapted to the specific language to improve performance further. The work by~\cite{vesely2012language,grezl2011study} presented bottleneck features for multi-lingual systems where they showed feature porting is possible and gave competitive results when compared to systems with mono-lingual features. Other approaches~\cite{tong2017investigation,vu2014multilingual} constructed a shared language independent phone set, which could then also be adapted to the target language. Our proposed model is inspired by the former approach which tries to learn latent features by sharing hidden layers across languages.


Connectionist Temporal Classification (CTC,~\cite{graves2006connectionist}) lends itself to low-resource multi-lingual experiments, because systems built on CTC tend to be significantly easier to train than those that have been trained using hidden Markov models~\cite{miao2015eesen,miao2016empirical}.~\cite{specom2017:mueller} shows that multi-lingual CTC systems with shared phones can improve performance in a limited data setting. As per our knowledge there has not been any prior work that have looked into learning ``bottleneck'' like features for a CTC based model and seen how it performs multi-lingually and cross-lingually with adaptation.

For this paper we use several languages from IARPA's Babel\footnote{This work used releases IARPA-babel105b-v0.4, IARPA-babel201b-v0.2b, IARPA-babel401b-v2.0b, IARPA-babel302b-v1.0a (these 4 languages will be called the ``MLing'' set), and IARPA-babel106b-v0.2g, IARPA-babel307b-v1.0b, IARPA-babel204b-v1.1b, IARPA-babel104b-v0.4bY (these 4 languages will be called the ``BAB300'' set), and IARPA-babel202b-v1.0d and IARPA-babel205b-v1.0 for testing.} project to test our model. These are mostly telephony (8kHz) conversational speech data in a low resource language. These were accompanied by a lexicon and dictionary in Extended Speech Assessment Methods Phonetic Alphabet (X-SAMPA) format. Table~\ref{tab:dataset} summarizes the amount of training data in hours along with the number of phonemes (including the CTC blank symbol) present for the languages we used in our experiments on the ``Full Language Pack'' (FLP) condition.

\begin{table}[h!]
\centering
\caption{Overview of the FLP Babel Corpora used in this work.}
\label{tab:dataset}
\begin{tabular}{|l|l|l|l|}
\hline
Subset  & Language  & \# Phones + $\emptyset$ & Training Data \\ \hline
        & Turkish   & 50                   & 79 hrs     \\
MLing   & Haitian   & 40                   & 67 hrs     \\
        & Kazakh    & 70                   & 39 hrs     \\
        & Mongolian & 61                   & 46 hrs     \\ \hline
        & Amharic   & 67                   & 43 hrs     \\
Bab300  & Tamil     & 41                   & 69 hrs     \\
        & Tagalog   & 48                   & 85 hrs     \\
        & Pashto    & 54                   & 78 hrs     \\ \hline
For     & Kurmanji  & 45                   & 42 hrs     \\
testing & Swahili   & 40                   & 44 hrs     \\ \hline
\end{tabular}
\end{table}

\begin{table*}[ht!]
\centering
\caption{Word (\% WER) and phoneme error rate (\% PER) for each of the test languages, on the Babel conversational development test sets.}
\label{tab:block-results}
\begin{tabular}{@{}ccccccccc@{}}
\toprule
Model                                    & \multicolumn{2}{c}{Kazakh}                         & \multicolumn{2}{c}{Turkish}                        & \multicolumn{2}{c}{Haitian}                        & \multicolumn{2}{c}{Mongolian} \\
                                         & WER           & PER                                & WER           & PER                                & WER           & PER                                & WER           & PER           \\ \midrule
\multicolumn{1}{c|}{Mono-lingual}         & 55.9          & \multicolumn{1}{c|}{40.9}          & 53.1          & \multicolumn{1}{c|}{36.2}          & 49.0          & \multicolumn{1}{c|}{36.9}          & 58.2          & 45.2          \\
\multicolumn{1}{c|}{Multi-lingual (MLing)}   & 53.2          & \multicolumn{1}{c|}{36.5}          & 52.8          & \multicolumn{1}{c|}{34.4}          & 47.8          & \multicolumn{1}{c|}{34.9}          & 55.9          & 41.1          \\
\multicolumn{1}{c|}{MLing \& FineTuning (FT)} & 50.6          & \multicolumn{1}{c|}{35.1}          & 49.0          & \multicolumn{1}{c|}{32.2}          & 46.6          & \multicolumn{1}{c|}{33.2}          & 53.4          & 39.6          \\
\multicolumn{1}{c|}{MLing + SWBD}           & 52.3          & \multicolumn{1}{c|}{36.6}          & 51.3          & \multicolumn{1}{c|}{33.0}          & 45.8          & \multicolumn{1}{c|}{33.9}          & 54.5          & 40.2          \\
\multicolumn{1}{c|}{MLing + SWBD \& FT}       & \textbf{48.2} & \multicolumn{1}{c|}{\textbf{33.5}} & \textbf{48.7} & \multicolumn{1}{c|}{\textbf{31.9}} & \textbf{44.3} & \multicolumn{1}{c|}{\textbf{31.9}} & \textbf{51.5} & \textbf{37.8} \\ \bottomrule
\end{tabular}
\end{table*}

\section{Multi-lingual CTC Model}
\label{sec:model}

A model trained with CTC loss is a sequence based model which automatically learns alignment between input and output by introducing an additional label called the blank symbol ($\emptyset$), which corresponds to `no output' prediction. Given a sequence of acoustic features $\bf{X} = (x_1,\dots, x_n)$ with the label sequence $\bf{z} = (z_1,\dots,z_u)$, the model tries to maximize the likelihood of all possible CTC paths $\bf{p} = (p_1,\dots,p_n)$ which lead to the correct label sequence $\bf{z}$ after reduction. A reduced CTC path is obtained by grouping the duplicates and removing the $\emptyset$ (e.g. $\mathcal{B}(AA \emptyset AABBC) = AABC$).
\begin{align*}
P(\mathbf{z}|\mathbf{X}) = \sum_{\mathbf{p} \in \texttt{CTC\_Path(\textbf{z})}} P(\mathbf{p}|\mathbf{X})
\end{align*}
Like in~\cite{miao2015eesen} we use this loss along with stacked Bidirectional LSTM layers to encode the acoustic information and make frame-wise predictions. \\
In our CTC multi-lingual model, we share the bidirectional LSTM encoding layer till the final layer and project the learned embedding layer to the phones of the respective target languages. The intuition behind this model is that training on more than one language will help in better regularization of weights and learning a better representation of features, as it will be trained on more data. We hypothesize that the final phoneme discrimination can be learned in a linear projection of the last layer. Figure~\ref{fig:speech_production} shows the schematic diagram of our multi-lingual model. Mathematically this can be written as,
\begin{align*}
&\mathbf{X} = \{\mathbf{X}_{L1} \cup \mathbf{X}_{L2} \cup \mathbf{X}_{L3} \dots \mathbf{X}_{Ln}\} && \mathbf{X}_{Li} = (x_{Li}^1,\dots, x_{Li}^n) \\
&\mathbf{e} = \texttt{Encoder}_{BiLSTM}(X) && \mathbf{e} \in \mathbb{R}^{n \times 2*h_{dim}}
\end{align*}
\begin{align*}
&P(\mathbf{p}|\mathbf{X}) = \begin{cases} 
      \texttt{softmax}(\mathbf{W}_{L1}\mathbf{e} + \mathbf{b}_{L1}) & \text{if } \mathbf{X} \in \mathbf{X}_{L1} \\
      \texttt{softmax}(\mathbf{W}_{L2}\mathbf{e} + \mathbf{b}_{L2}) & \text{if } \mathbf{X} \in \mathbf{X}_{L2} \\
      \dots \\
      \texttt{softmax}(\mathbf{W}_{Ln}\mathbf{e} + \mathbf{b}_{Ln}) & \text{if }\mathbf{X} \in \mathbf{X}_{Ln}
\end{cases}
\end{align*}

Unlike~\cite{vesely2012language}, we do not have any bottleneck layer, and the whole model is sequence trained based on CTC loss. Note that here we recognize a sequence of phonemes which is a much harder problem. Traditional HMM/DNN systems perform frame-wise recognition of individual phonemes, usually relying on alignments that have been generated by mono-lingual models. This can be considered a much simpler task than the recognition of a phone sequence.

\section{Experiments and Observations}
\label{sec:experiments}
\subsection{Multi-lingual CTC model}
To align with project goals, we chose to perform experiments on a set of four languages which are the closest/ have maximum phone overlap with Kurmanji -- Kazakh, Turkish, Mongolian and Haitian.
We used a 6-layer bidirectional LSTM network with 360 cells in each direction, which performed best on average across the majority of Babel languages in a systematic search experiment. Table~\ref{tab:block-results} shows the results. For consistency, we used absolutely identical settings across all languages, and did not perform any language-specific tuning, other than choosing the lowest perplexity language model between 3-gram and 4-gram models for WFST-based decoding. Techniques such as blank scaling and applying a softmax temperature can often improve results significantly, but we did not apply any of them here for consistency.

In our multi-lingual experiments, we use the same 6-layer Bi-LSTM network with 360 cells (per direction) in each layer as our shared encoded representation\footnote{The code to train the multi-lingual model will be released as part of EESEN~\cite{miao2015eesen}.}. Again, this setup performed best on average on a larger set of languages. Multi-lingual training on the ``MLing'' set (the four languages shown in Table~\ref{tab:block-results}) improves WER by 1.7\% (absolute) on average, while keeping the LSTM layers shared across all languages. If we fine-tune the entire model towards each language specifically, performance improves further, by 4.4\% on average over the baseline. If we roughly double the amount of training data by adding the Switchboard 300h training set to the ``MLing'' training data, performance improves yet again, for both the universal (MLing+SBWD) and language-specific (MLing+SWBD \& FT) case. Overall, WER and PER improve by about 6\% absolute ($>$10\% relative), which is in line with other results reported on comparable tasks discussed in section~\ref{sec:related}.

As expected, reductions in the error rates tend to be higher for the lower resource languages, like Kazakh and Mongolian. 

\subsection{Data Selection}

Given that adding a seemingly unrelated, but high resource language improved the performance of the model on four low resource languages, we further studied the impact of varying the source(s) of the extra data. Specifically, we replaced the 300\,h Switchboard corpus with four more unrelated Babel languages, ``BAB300'' composed of Tamil, Amharic, Pashto, and Tagalog.
The results on the test data are summarized in Table~\ref{tab:bab300}. We can see that adding Switchboard data outperforms adding more unrelated Babel languages.
\begin{table}[H]
\centering
\caption{Word error rate (\% WER) on the test languages when switching the SWBD data with 300 hrs equivalent of Babel.}
\label{tab:bab300}
\resizebox{\linewidth}{!}{
\begin{tabular}{ccccc}
\hline
Model   & Kazakh        & Turkish       & Haitian       & Mongolian     \\ \hline
MLing + BAB300 & 57.5          & 52.0          & 47.8          & 56.7          \\
MLing + SWBD   & \textbf{52.3} & \textbf{51.3} & \textbf{45.8} & \textbf{54.5} \\ \hline
\end{tabular}}
\end{table}

While our main goal here has been the creation of a multi-lingual recognizer, we verified that models that have been trained on a single Babel language plus 300\,h of Switchboard do not outperform the fine-tuned MLing+SWBD system, while there is no clear pattern on other languages. This indicates that it is generally beneficial to train (sequence-based) multi-lingual systems on closely related languages, and/or on large amounts of well-prepared but unrelated mono-lingual data, but that adding a large number of languages may in fact prevent the model from training well.

\subsection{Representation Learning}

\begin{figure}[ht!]
  \centering
  \includegraphics[width=\linewidth]{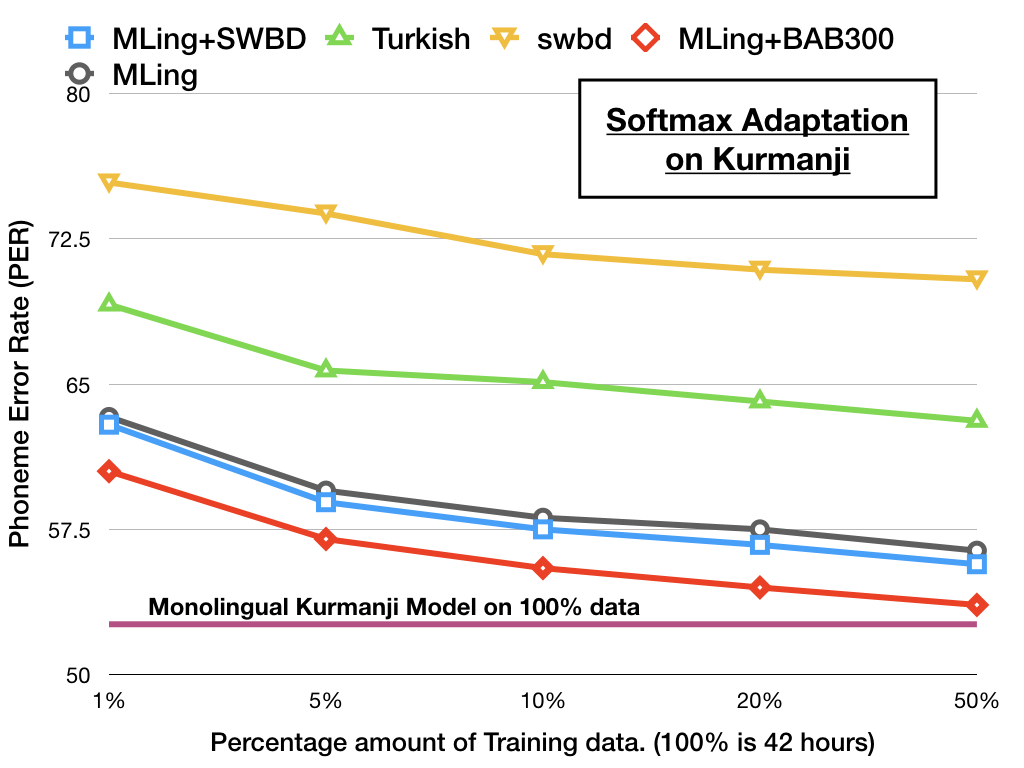}
  \caption{Cross-lingual training of CTC softmax layer only on top of different ``donor'' models.}
  \label{fig:cross}
\end{figure}

In order to study to what extent the CTC sequence models have learned useful bottleneck like discriminatory audio features that are independent of the input language, we attempt to port a model to an unseen language. We aim to use the trained model as a language-independent feature extractor that can linearly separate any language into a phoneme sequence. To do this, we replace the softmax layer (or ``layers'' in the multi-lingual case) of a ``donor'' CTC model with a single softmax, which we then train with varying amounts of data from the target language, Kurmanji in our case. 
Figure~\ref{fig:cross} shows how different ``donor'' models behave in this situation. In the cross-lingual case, it becomes beneficial to train the LSTM layers with as many different languages as possible (``MLing+BAB300'' outperforms ``MLing+SWBD'' and ``MLing''), while a single related language (Turkish) outperforms adaptation on a larger amount of data from an unrelated language (SWBD). There is a large gap between mono-lingual systems and multi-lingual systems. Improvements become smaller once training is performed on 4\,h (10\%) of data or more, but even then the re-estimation of the softmax layer (with ca.~32k parameters) benefits from more data.

\begin{figure*}[ht!]
  \centering
  \subfloat[][Adaptation of softmax layer only for Kurmanji and Swahili\\
  targets. Kurmanji performs well, because the language is similar to\\ some training languages.]{
  \includegraphics[width=0.465\linewidth]{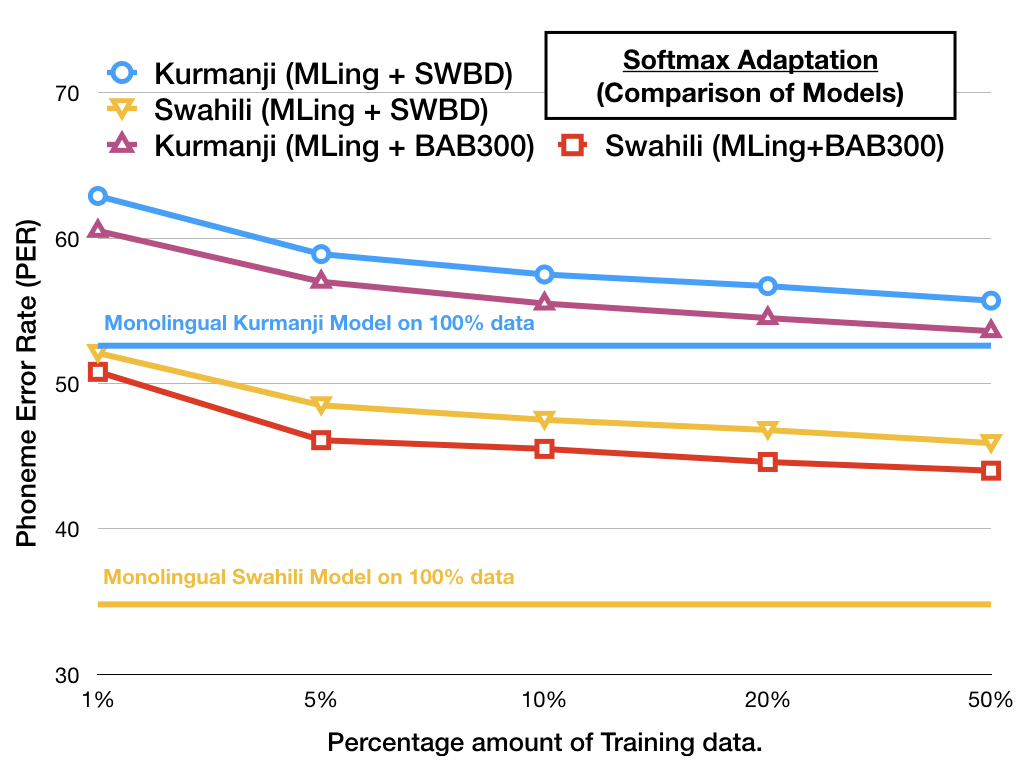}}
  \subfloat[][Adaptation of entire network (re-training) to target languages.\\
  This outperforms softmax adaptation (on the left) as soon as 2-4\,h of data become available.]{
  \includegraphics[width=0.465\linewidth]{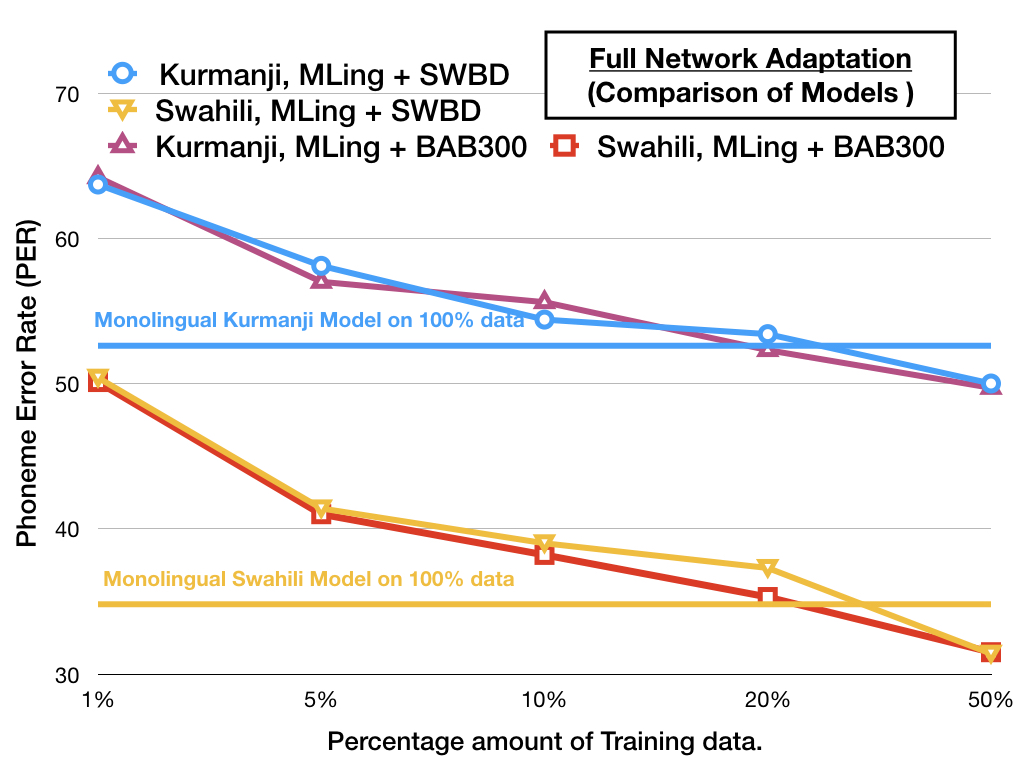}}
  \caption{Cross-lingual training of Kurmanji and Swahili systems.}
  \label{fig:cross_ft1}
\end{figure*}

It thus seems that multi-lingual systems do indeed learn a portable, language independent representation, which is useful when porting to a new language, while the sheer amount of data is less beneficial.

\subsection{Cross-lingual Explorations}

\begin{figure}[hb!]
  \centering
  \includegraphics[width=\linewidth]{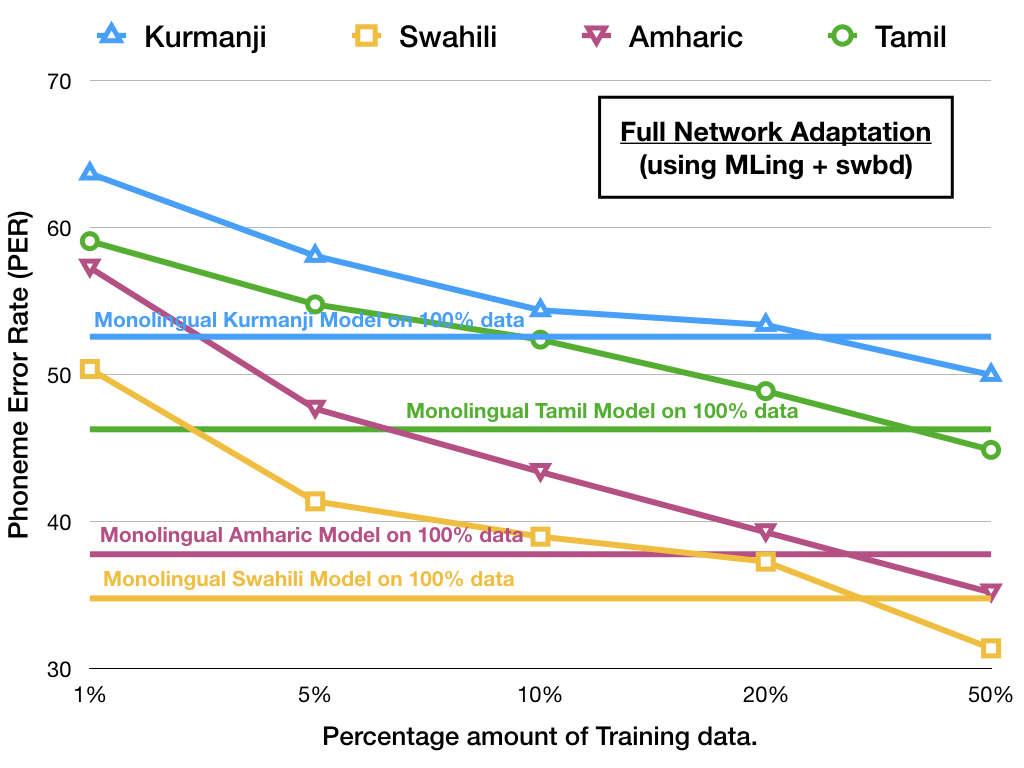}
  \caption{PER on different amounts of cross-lingual data using a full network end-to-end adaptation (retraining).}
  \label{fig:cross_ft}
\end{figure}

Figure~\ref{fig:cross_ft} shows that for both related and unrelated languages, a multi-lingual system surpasses the mono-lingual baseline once about 25\% of the original data has been seen. The behavior of retraining (``full network adaptation'') seems independent of the original trained languages.

To further investigate how multi-lingual models can be used in cross-lingual settings, and with varying amounts of training data, we compare ``softmax'' adaptation and full network adaptation (retraining) on Kurmanji and Swahili, two languages which we did not see in training.
We use the (MLing + SWBD) and (MLing + BAB300) ``donor'' models. Figure~\ref{fig:cross_ft1} shows that for small amounts of adaptation data, and a target language that is related to the pre-trained languages (Kurmanji), ``softmax adaptation'' is competitive, and an initialization with many languages is beneficial.

When the entire network can be retrained (``full network adaptation'', shown on the right side of Figure~\ref{fig:cross_ft1}), there is very little difference between the ``donor'' systems' performance.



\section{Conclusion}
\label{sec:con}

In this paper, we demonstrate that it is possible to train multi-lingual and cross-lingual acoustic models directly on phone sequences, rather than frame-level state labels. Unlike multi-lingual bottleneck features, these CTC models do not require the generation of state alignments, which facilitates their use.

In multi-lingual settings, it seems beneficial to train on related languages only, or on large amounts of clean data; there is no benefit simply from training on many languages. It is thus possible to combine e.g.~Switchboard and Babel data.

In very low resource cross-lingual scenarios, it is possible to adapt a model to a previously unseen language by re-training the softmax layer only. CTC models can learn a language independent representation at the input to the softmax layer. We find that training the models trained on related languages help, as does training on many languages, rather than large amounts of data. As more and more data is available, and the whole network can be retrained, and the effect of the choice of language for the multi-lingual training disappears.

As future work, we are investigating on decoding the CTC output using a phoneme based neural language models trained on non-parallel text, thereby facilitating us to do zero-resource speech recognition. 


\section{Acknowledgements}
\label{sec:ack}

This project was sponsored by the Defense Advanced Research Projects Agency (DARPA) Information Innovation Office (I2O), program: Low Resource Languages for Emergent Incidents\\ (LORELEI), issued by DARPA/I2O under Contract No. HR0011-15-C-0114. 

We gratefully acknowledge the support of NVIDIA Corporation with the donation of the Titan X Pascal GPU used for this research. 
This work used the Extreme Science and Engineering Discovery Environment (XSEDE), which is supported by National Science Foundation grant number OCI-1053575.  Specifically, it used the Bridges system, which is supported by NSF award number ACI-1445606, at the Pittsburgh Supercomputing Center (PSC).

We are grateful to Anant Subramanian and Soumya Wadhwa for their feedback on the presentation of this paper.

\bibliographystyle{IEEEbib}
\bibliography{refs.bib}

\end{document}